\documentclass[twoside,11pt]{article}
\usepackage{akbc}
\akbcheading
\ShortHeadings{Gollum: A Gold Standard for Multi Source Knowledge Graph Matching}{Hertling \& Paulheim}
\finalcopy

\usepackage{graphicx}
\usepackage{multirow}
\usepackage{hyperref}

\usepackage{adjustbox}

\begin{document}
\title{Gollum: A Gold Standard for Large Scale\\Multi Source Knowledge Graph Matching}

\author{\name Sven Hertling \email sven.hertling@uni-mannheim.de \\
       \addr Data and Web Science Group, University of Mannheim, Germany
       \AND
       \name Heiko Paulheim \email  heiko.paulheim@uni-mannheim.de \\
       \addr Data and Web Science Group, University of Mannheim, Germany}

\maketitle

\begin{abstract}

The number of Knowledge Graphs (KGs) generated with automatic and manual approaches is constantly growing.
For an integrated view and usage, an alignment between these KGs is necessary on the schema as well as instance level.
While there are approaches that try to tackle this multi source knowledge graph matching problem,
large gold standards are missing to evaluate their effectiveness and scalability.
We close this gap by presenting \textbf{Gollum} -- a \textbf{gol}d standard for \textbf{l}arge-scale m\textbf{u}lti source knowledge graph \textbf{m}atching with over 275,000 correspondences between 4,149 different KGs.
They originate from knowledge graphs derived by applying the DBpedia extraction framework to a large wiki farm.
Three variations of the gold standard are made available:
(1) a version with all correspondences for evaluating unsupervised matching approaches, and two versions for evaluating supervised matching: (2) one where each KG is contained both in the train and test set, and (3) one where each KG is exclusively contained in the train or the test set.

\end{abstract}

\section{Introduction}
Knowledge graphs have turned into a versatile and universal data representation mechanism and can be used in many data-intensive applications.
While many applications require information from different knowledge graphs, the parallel use of multiple knowledge graphs often requires integrating the data from those graphs into a common graph first. To that end, matching both the schema and the instances of the knowledge graph is required.
To solve that challenge, different knowledge graph matching tools have been proposed. 
One limitation is that those tools usually take only a pair of knowledge graphs as input. In contrast, there are quite a few scenarios where the number of knowledge graphs is larger. This is also the case for knowledge graph construction from diverse sources, where one possible approach is to first create an isolated knowledge graph from each source, and then integrate those graphs later. This requires solutions which can handle the matching of multiple knowledge graphs.

So far, existing tools, as well as gold standards, are designed in a pairwise fashion. 
Since there can be vicious circles of available gold standards and tools (i.e., tools are less likely to be developed for a task for which there is no evaluation dataset), we propose \emph{Gollum}, a gold standard for large-scale multi-source knowledge graph matching. 
We describe the creation of the gold standard and present an evaluation of its quality.

\section{Related Work}
\label{sec:related}

In this section, we present and compare existing gold standards. We categorize them into entity matching and KG matching, and into 1:1 vs. multi source datasets.

In entity matching, the schema is fixed and the same across all sources while, at the same time, the number of attributes is quite small. 
This means that the matching task boils down to comparing the attributes of the entities.
In comparison, knowledge graphs allow storing instances of different classes which might need different features/rules for matching.
Furthermore, the properties (attributes) are also not homogenized.
This makes the task even more difficult but at the same time more realistic.
Further, both flavors can exist as $1:1$ or multi source problems. In the latter case, more than two data sources are to be matched.

	\begin{table}[t]
		\centering
		\caption{Comparison of related work}
		\label{tab:relwork}
		\resizebox{\textwidth}{!}{%
			\begin{tabular}{|c|l|l|r|r|r|r|r|l|}
				\hline
				\multicolumn{1}{|l|}{\textbf{}} &
				\textbf{} &
				\textbf{Dataset} &
				\textbf{\#sources} &
				\textbf{\#entities} &
				\textbf{\#attr} &
				\textbf{\#matches} &
				\textbf{\#clusters} &
				\textbf{generator} \\ \hline
				\multirow{9}{*}{\begin{tabular}[c]{@{}c@{}}entity\\ matching\end{tabular}} &
				\multirow{4}{*}{1:1} &
				DBLP-ACM &
				2 &
				4,908 &
				4 &
				2,224 &
				- &
				no \\ \cline{3-9} 
				&
				&
				DBLP-Scholar &
				2 &
				66,879 &
				4 &
				5,347 &
				- &
				no \\ \cline{3-9} 
				&
				&
				Amazon-Google &
				2 &
				4,589 &
				4 &
				1,300 &
				- &
				no \\ \cline{3-9} 
				&
				&
				Abt-Buy &
				2 &
				2,173 &
				4 &
				1,097 &
				- &
				no \\ \cline{2-9} 
				&
				\multirow{4}{*}{\begin{tabular}[c]{@{}l@{}}multi\\ source\end{tabular}} &
				Affiliations &
				1 &
				2,260 &
				1 &
				32,816 &
				330 &
				no \\ \cline{3-9} 
				&
				&
				Geographic &
				4 &
				3,054 &
				3 &
				4,391 &
				820 &
				no \\ \cline{3-9} 
				&
				&
				Music Brainz 20K &
				5 &
				19,375 &
				5 &
				16,250 &
				10,000 &
				yes \\ \cline{3-9} 
				&
				&
				\begin{tabular}[c]{@{}l@{}}North Carolina\\ Voters 5M\end{tabular} &
				5 &
				5,000,000 &
				4 &
				3,331,384 &
				3,500,840 &
				yes \\ \cline{3-9} 
				&
				&
				WDC - LSPM &
				79,123 &
				26,000,000 &
				10 &
				40,582,671 &
				16,391,439 &
				no \\ \hline
				\multirow{3}{*}{\begin{tabular}[c]{@{}c@{}}KG\\ matching\end{tabular}} &
				\multirow{2}{*}{1:1} &
				DB to WD &
				2 & 57,296,772
				&
				- &
				376,065 &
				- &
				no \\ \cline{3-9} 
				&
				&
				OAEI KG track &
				8 &
				493,845 &
				- &
				15,359 &
				- &
				no \\ \cline{2-9}
				&
				\begin{tabular}[c]{@{}l@{}}multi\\ source\end{tabular} &
				\textbf{Gollum} 
				& 4,149
				& 10,704,943
				& -
				& 275,103
				& 87,926
				& 
				no \\ \hline
			\end{tabular}
		}
	\end{table}

Table~\ref{tab:relwork} shows the overview of datasets.
The most prominent datasets are 1:1 entity matching datasets such as DBLP-ACM, DBLP-Scholar, Amazon-Google, and Abt-Buy.
The first two are initially presented in \cite{kopcke2008training} and originate from the bibliographic domain. The two E-commerce datasets are used in \cite{kopcke2010evaluation} to evaluate their entity resolution system.

Multi source entity matching datasets are more closely related to Gollum. Here, the schema is still the same across all datasets, but the matching system needs to deal with more than two sources.
The affiliations dataset\footnote{\url{https://dbs.uni-leipzig.de/research/projects/object\_matching/benchmark\_datasets\_for\_entity\_resolution}} is an exception because it only contains one source but this requires a clustering of entities (thus it is closer to a multi source scenario). It
consists of only one attribute which are affilations strings extracted from publications (entities need to be clustered). The geographic dataset~\cite{saeedi2017comparative} contains real-world entities from DBpedia, Geonames, Freebase, and NYTimes. It has been used at the instance matching track of OAEI 2011\footnote{\url{http://oaei.ontologymatching.org/2011/instance/}}.
The Music Brainz and North Carolina Voters datasets~\cite{saeedi2017comparative} are based on records about songs and persons. Those datasets are corrupted by tools like the DAPO data generator~\cite{hildebrandt2017large} and GeCo~\cite{christen2013flexible}. 
Thus these datasets can be arbitrarily large but do not represent real-world matching problems. The WDC-LSPM dataset \cite{primpeli2019wdc} is about product matching.
It is created by extracting structured data from webpages (e.g. microdata and RDFa). They focus on products because they contain identifiers that are shared among those web pages. Thus a high number of sources and matches can be achieved.
\cite{zhao2020experimental} and \cite{leone2022critical} provide surveys about entity alignment approaches together with datasets on which they are evaluated.

In contrast, knowledge graph matching datasets only exist as $1:1$ problems. 
\cite{azmy2019matching} presents a gold standard between DBpedia and Wikidata to evaluate their embedding based entity matching approach.
Their main focus is on disambiguating entities and thus they only use entities that appear multiple times with a similar label, e.g. a person with the name ``John Burt'' (can be the footballer, the field hockey player, etc.). The OAEI KG track~\cite{hertling2020knowledge} contains only eight KGs in a one to one matching scenario.

To the best of our knowledge, no further multi source KG matching datasets exist. The work presented in this paper aims at closing that gap.

\section{Generating the Gold Standard}
\label{sec:approach}
There are many ways of generating a gold standard for a multi source matching task.
In \cite{hertling2020knowledge}, we have shown that a crowd-sourced approach cannot yield a lot of \textit{interesting} correspondences, i.e., correspondences between entities that are very similar but not the same (candidates for false positives)
or very dissimilar but actually referring to the same entity (e.g. with different names or labels), and, at the same time, that it is very expensive to create a large-scale, high-quality gold standard.
Therefore, we pursue a different approach in this paper. 
We create knowledge graphs from Wikis of Web pages representing entities and collect links between those Web pages. Then, we subsequently filter and enhance that set so that it contains mostly equivalence links.

\subsection{Knowledge Graphs and Alignments}
\begin{figure}[t]
	\centering
	\includegraphics[width=0.9\textwidth]{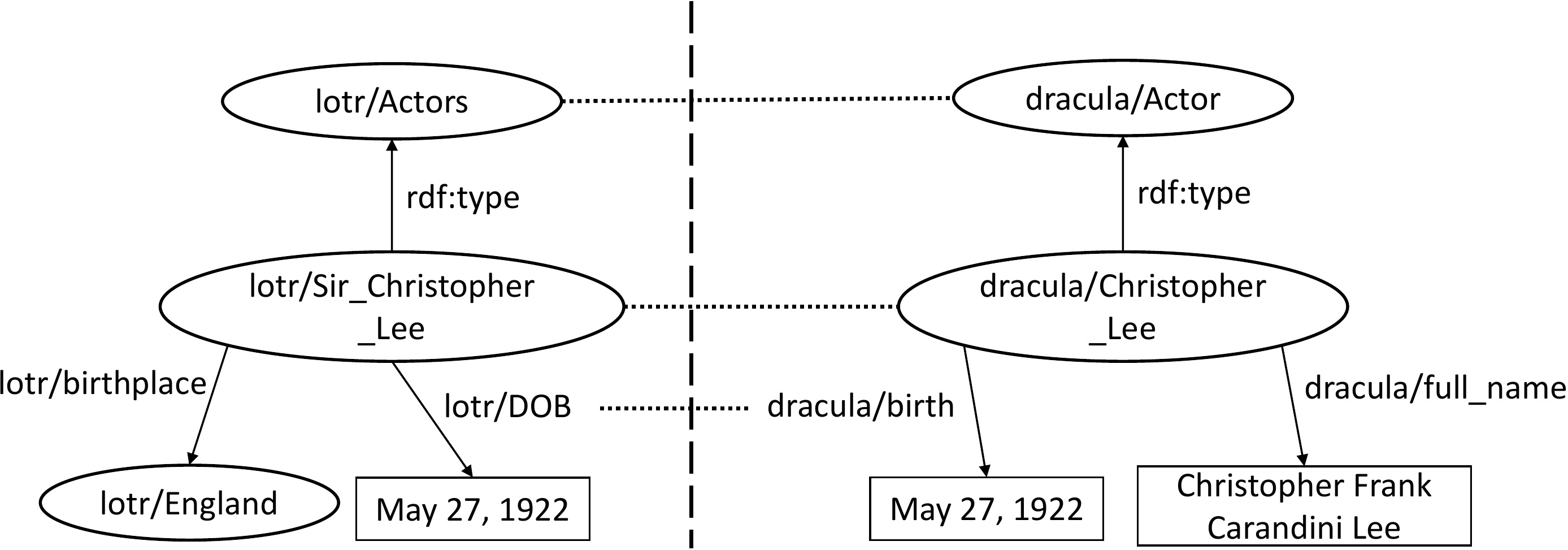}
	\caption{Example of resources exrtacted from the Lord of the Rings Wiki (left) and the Dracula Wiki (right). Each wiki page is transformed to a KG resource and corresponding templates are used to create classes. Correspondences may exist between classes, properties, and instances.}
	\label{fig:example}
\end{figure}

The KGs which are used in this work are generated by applying the DBpedia extraction framework \cite{auer2007dbpedia} to Wikis which are served by the MediaWiki framework. Wikiapiary\footnote{\url{https://wikiapiary.com}} lists such Wikis and also includes so-called \emph{wiki farms} (wiki-families). One prominent example is Fandom\footnote{\url{https://www.fandom.com} (formerly wikia.com)}. It allows the creation of collaborative Wikis for any topic. Those Wikis are crawled and processed by the DBpedia extraction framework. This framework is specially designed for Wikipedia and thus it needs to be modified to work with Wikis using other conventions, e.g. for guaranteeing stable link and abstract extraction. The extraction resembles the creation of DBpedia: each wiki page is transformed to a KG instance, and all templates (especially the infobox template) are used to extract properties and corresponding values. The name of the templates is further used to derive the classes. An example is shown in figure \ref{fig:example}.

With this approach, 307,466 KGs are generated which contain complementary information to Wikipedia/DBpedia. Thus more information about less famous entities (also called \emph{long-tail entities}) is included in these wikis. The disadvantage is that many Wikis contain the same entities and schemas which need to be aligned to create a consolidated KG. An alignment between those KGs is defined as a set of correspondences like $<A, B, =, c>$ where the first and second element represents the URI of the resources. The third component defines the relation which holds between those resources (here, we only consider equivalence). The last component is the confidence $c \in \left[0;1\right]$.
In the next section the wiki pages are analyzed to extract \texttt{sameas} links.

\subsection{Extracting Candidate Links}
To identify pages denoting the same entity, we look for specific sections on those pages that contain links to the same entities in other wikis. Typical titles for such sections are 'external links', 'trivia', 'weblinks', or 'see also'. This section title varies between Wikis but is usually used consistently \textit{within} a wiki.
In all subsequent steps, only links which point to another wiki page are used (so-called inter wiki links).
The goal of this step is to identify those sections which contain links to same entities in other wikis.

First, all links which point to a page with the \emph{same title} in another Wiki as the source page are extracted. 
Those correspondences are assumed to link same entities.
The section titles of those links are extracted to find the best title which represents a section containing \texttt{same as} links.
It is later also used to extract links that do not share the same title (so called hard positives).

Given the section titles, the goal is to extract a rather large substring that covers at the same time most of the given links.
Thus is is feasible to cover e.g. 'external links' and 'external link' at the same time.
All possible substrings of the section titles are computed which are at least of length two. Afterward, we count how often each substring covers a section title.
Short strings cover a lot of titles, but longer substrings are preferable. Thus the length of the title as well as the number of covered links is normalized to the range between zero and one (diving by the longest title or respectively the maximum number of links). Due to the fact that both values should be high, the harmonic mean of both values is used to calculate the final quality $Q$ which is defined as:
\begin{equation}
	Q(text) = \frac{
		2 * \frac{coveredTitles(text)}{\#links} * \frac{length(text)}{length(longestTitle)}
	}{
	        \frac{coveredTitles(text)}{\#links} + \frac{length(text)}{length(longestTitle)}
    }
\end{equation}

The title substring with the highest quality value $Q$ is used to extract all links in sections that contain this substring.
This approach is only applied if (1) at least five links exist in sections sharing the same title and (2) the fraction of extracted links that share the same title is greater than 20\%. In all other cases, the default section title ''link'' is used.

The resulting alignment still contains many correspondences which do not represent the same entity, but rather related entities. Therefore, we perform further steps to refine the alignment.

\subsection{Normalization and Injectivity}

\begin{figure}[t]
	\centering
	\includegraphics[width=0.5\textwidth]{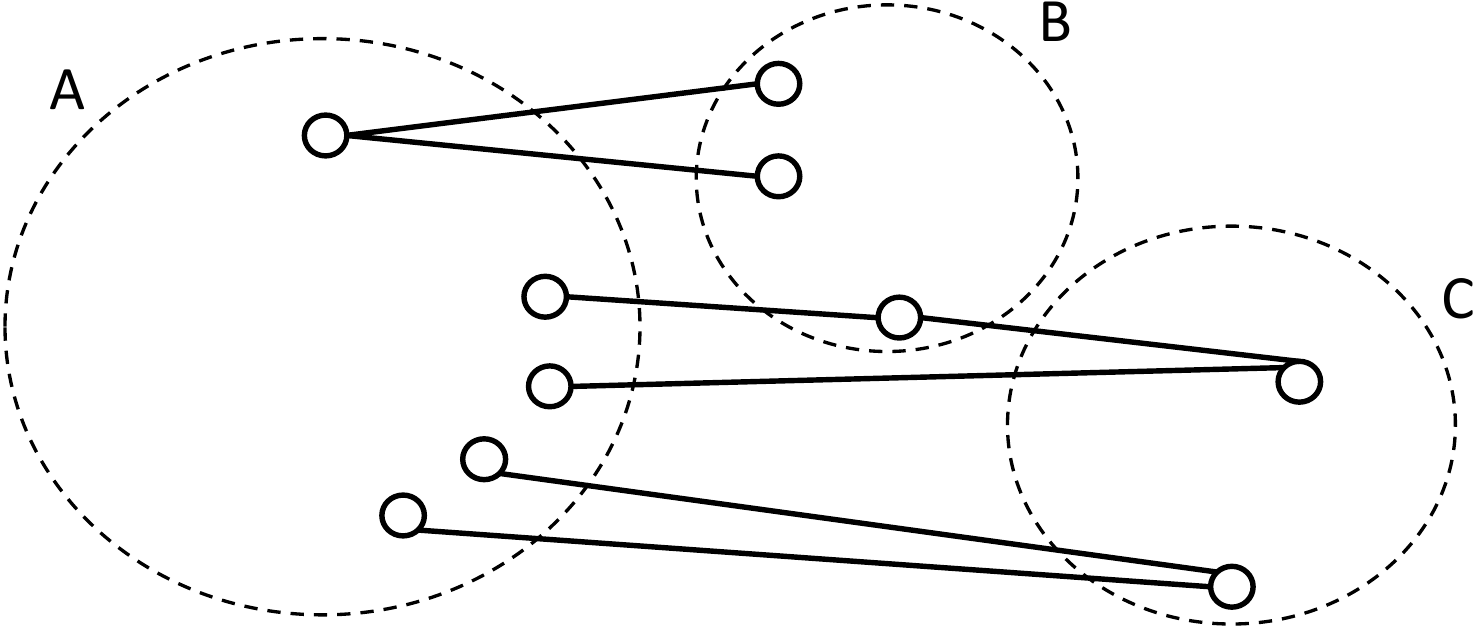}
	\caption{Example of links between Wikis A, B, and C. The upper and lower two can be removed as well as one of the three links in the middle.}
	\label{fig:links}
\end{figure}

As a very first step, the direction of the correspondences is normalized.
We compare the name of the wiki, to define one canonical direction. The confidence is set to 0.5 if the link only exists in one direction, and to 1.0 for bidirectional connections. This value will be further used to remove wrong correspondences.
In addition, all redirects are resolved.

We assume that each wiki contains only one page for each entity, and we use that assumption to remove further invalid links.
Figure \ref{fig:links} shows an example.
The dotted circles are Wikis named A, B, and C. The upper two correspondences are removed because an entity in A links to two entities in B which contradicts the assumption. One of these links might be correct but it is not easily possible to tell which one it is. To maximize the precision of the gold standard, all of these links are removed. 
The lower two correspondences between A and C are removed using the same rationale.

The three links in the middle do also not satisfy the assumption (at least when considering the transitivity of the equivalence relation). This case is discussed in section \ref{sec:transitiveClosure}.

\subsection{Removing Links to Disambiguation Pages and Dead Links}

After an analysis of the resulting alignment, it turned out that some pages are disambiguation pages that should not be a part of the final alignment and whenever a source or target of a correspondence is a disambiguation page, it should be removed.
The DBpedia extraction framework detects such pages by searching for Wikimedia templates which contain the text \texttt{Disambig} and extracts all links which point to the disambiguated articles.
This works fine in a controlled scenario like Wikipedia where articles use a predefined set of templates. 
In wiki farms such as Fandom this strict usage of specific names is not present.

Thus, a more robust disambiguation detection is needed.
Three approaches are implemented. They are listed here in the order of quality:  1) the text \texttt{disambiguation} appears in the label (title) of the article 2) one of the following variants of text appears in the first sentence of the article: \texttt{\{can, could, may, might\} refer to} 3) the article appears in a category which contains the text \texttt{disambiguation}. 

To further clean up the alignment, so-called anchor links (links containing a hash sign '\#') are removed as well. They refer to a part of a page and not to the whole page.
These parts of the pages do not represent an entity in the corresponding KG and can thus not be used in the alignment.
Afterward, all pages which do not appear in the KG are removed as well. This usually corresponds to the situation where a page links to a nonexistent 
or already deleted page (dead link). Furthermore, it is ensured that the entity can actually be found in the KG, i.e., it appears as a subject or object in a triple.

\subsection{Transitive Closure}
\label{sec:transitiveClosure}

\begin{figure}[t]
	\centering
	\includegraphics[width=0.8\textwidth]{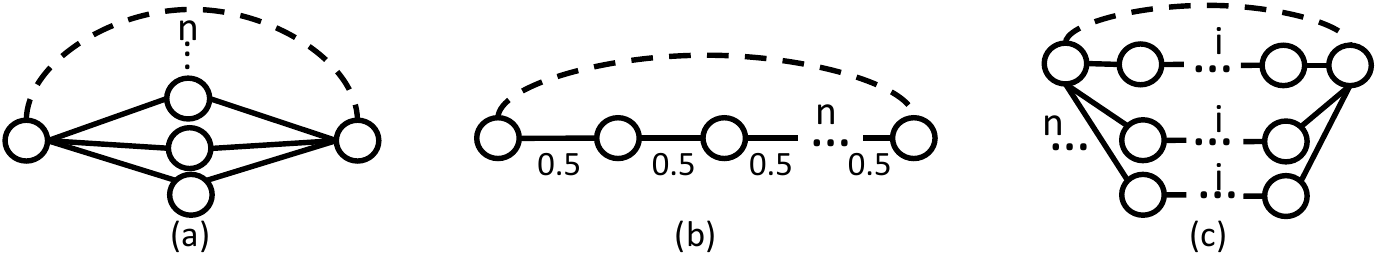}
	\caption{The confidence of the dotted edge (transitive edge) should be determined. (a) shows the best case, (b) the worst case, and (c) a case where the shortest path is long and the flow is high.}
	\label{fig:trans}
\end{figure}

In the next refinement step, the transitivity of the equivalence relation is used to remove additional correspondences.
First, the transitive closure of the whole alignment is calculated (implementation based on \cite{beek2018sameas} (Section 3.3)).
Afterward, each identity set is analyzed if two entities of the same wiki are contained (see also the middle part of figure \ref{fig:links}).
This would violate the basic assumption that no real-world concept is represented multiple times in a wiki.
Thus at least one of these correspondences which forms this identity set is wrong and should be removed.
The alignment is sorted by criteria described below and correspondences are removed as long as the identity set contains two entities in the same wiki.

The first sorting criterion checks if the identity set could be repaired by just removing one correspondence. Those correspondences are sorted higher than those which would not.
The second criterion is based on the confidence defined earlier. Correspondences with a confidence of 1.0 (where the entities link to each other) are sorted lower than the ones with a confidence of 0.5. To break ties, the algorithm of \cite{raad2018detecting} is applied which scales well to large identity sets and computes an error value of each link based on graph metrics like community detection. Links with a higher error value are removed first. In case these values are not sufficient to differentiate each correspondence, the levenshtein distance between the labels is used. To make the ordering deterministic, we finally order by the entity URI.
Based on these criteria, all identity sets are resolved such that there are no entities of the same wiki under the transitive closure by removing correspondences following the sorting order. 
Similar approaches are also used in \cite{de2009towards}.

The transitive closure cannot only be used for removing, but also for adding correspondences.
For example, for links between entities $e_1$ and $e_2$ and $e_2$ and $e_3$ in three knowledge graphs $KG_1$, $KG_2$, and $KG_3$, we also want to include a link between $e_1$ and $e_3$ in our gold standard.
To that end, we also need to set a confidence value for the newly generated links.
Figure \ref{fig:trans} shows examples of correspondences to be added (dotted edges) for which the confidence needs to be determined. The optimal case is represented by (a):
There are a lot of connections between the source and target which indicates that the confidence should be set to a high value. In case the alignment is considered as a graph, this corresponds to a high flow between the two vertices. Thus the maximum flow is computed using Edmonds-Karp algorithm~\cite{edmonds1972theoretical}. Using the flow as the only measure can fail in case (c) where the flow is high but the paths are rather long (large $i$).
In general shorter paths are preferable to longer ones. In order to penalize paths through low-confidence edges, we compute the path weight by summing up the terms $1.5-c_i$ for each confidence $c_i$ of an edge in the path.\footnote{Essentially, this flips the 0.5 and 1.0 confidence.}
Case (b) represents the worst case which has a rather low maximum flow and the path between the entities is rather long.
For all graph-related algorithms, the implementations of JGraphT~\cite{jgrapht} are used.

To compute a confidence, given the length of the shortest path and the maximum flow, we use the harmonic mean of $1-\frac{1}{maxFlow + 1}$ and $\frac{1}{shortestPathLength}$. The reason for using these formulas is that the flow value can be between 0.5 and infinity (a high value of n in figure \ref{fig:trans}(a)) and a high flow value should result in a confidence close to one. 
For the shortest path, a long path should result in a low confidence. Both values should be high to indicate a good connection between the entities, which is enforced by using the harmonic mean.
The confidence of correspondences generated under the transitive closure is between zero and one. To separate the direct extracted links from the transitive links, the latter ones are stored in an extra file.

\subsection{Schema Matches}

All correspondences extracted so far are only on the instance level but the KGs also contain classes and properties which are not aligned.
Therefore in the next step, all instance correspondences are used to extract a schema alignment.

The underlying idea is that two aligned classes also share at least a subset of their instances. Thus, for each combination of classes in two KGs, 
the number of shared instances is computed.
Together with the number of instances for each class, it is possible to compute how likely the classes also match.
We experimented with two metrics: $Sim_{Dice}$\cite{diceSim} and $Sim_{Min}$\cite{kirsten2007instance} which are defined as 
\begin{equation}
Sim_{DICE}(c_1, c_2) = \frac{2 * \vert I_{c_1} \cap I_{c_2}\vert}{\vert I_{c_1} \vert + \vert I_{c_2} \vert} \in [0...1]
\end{equation}
\begin{equation}
Sim_{MIN}(c_1, c_2) = \frac{\vert I_{c_1} \cap I_{c_2}\vert}{min(\vert I_{c_1} \vert, \vert I_{c_2} \vert)} \in [0...1]
\end{equation}
where $I_{c_j}$ denotes the set of instances for class $c_j$. $I_{c_1} \cap I_{c_2}$ corresponds to the shared instances of $c_1$ and $c_2$.

Property matches are determined in the same way by just replacing $\vert I_{c_i} \vert$ with $\vert I_{p_i} \vert$ which denotes the number of occurences of property $p_i$ and $\vert I_{p_1} \cap I_{p_2} \vert$ denotes the number of occurences of two properties $p_1$ and $p_2$ with the same subject and object.

\subsection{Train/Test Split and Datasets}
 Many matching systems use supervised learning, e.g., to adapt their thresholds and weights. This is useful to generalize because each task might require different matching strategies and attributes to use. To account for those systems, multiple versions of this gold standard are released: (1) an unsupervised set that contains all correspondences (2) a supervised version where entities from each KG appear in the train and test set, and (3) a supervised version where all entities from each KG appear \emph{either} in the train or in the test set. For the latter two, an 80/20 split is used.

The test set for both supervised versions is optimized to contain none (or at least fewer) correspondences which could be inferred by computing the transitive closure of the training set. This is easier to achieve for version (2) than (3) because in (3) many KGs share at least one entity in the transitive closure.
Optimizing this requirement is achieved with the \texttt{Group\-Shuffle\-Split}\footnote{\url{https://scikit-learn.org/stable/modules/generated/sklearn.model_selection.GroupShuffleSplit.html}} class of scikit learn which allows specifying groups of examples which should be placed either in train or test.
For version (2) the groups correspond to the identity sets computed during the transitive closure. For version (3) the groups correspond to KGs. Those which share a lot of correspondences are placed in the same group.

 Up to now, only the gold standard is split. On the other hand, the KGs which serve as the input for the matching task are also available in different subsets.
 In the first subset (gold), only the KGs which actually participate in the gold standard are collected.
 This results in 4,149 KGs. The next larger dataset (40K) contains the largest 40,000 English KGs. Finally, the dataset (all) with 307,466 KGs is provided. The latter can be used to check for the scalability of multi source KG matching approaches.
 
 In the datasets, the links between pages are available during training and testing but not the actual structure of the pages including sections etc which are used to generate the gold standard.
 
 The datasets and gold standards are published in zenodo~\cite{zenodoDataset} 
 under a CC BY-SA license\footnote{\url{http://creativecommons.org/licenses/by-sa/3.0/}} due to the license of the original source\footnote{\url{https://www.fandom.com/licensing}}.

\section{Analysis of the Gold Standard}
\label{sec:analysis}
\begin{table}[t]
\caption{Overview of the alignment size after each step.}
\label{tab:eval}
\begin{adjustbox}{max width=\textwidth}
\begin{tabular}{|l|r|r|r|r|}
\hline
\textbf{Step} & \textbf{\#Links} & \textbf{\#Trivial} & \textbf{\#Non-trivial} & \textbf{\#KG Pairs} \\ \hline
1) Extracting Candidate Links      & 218,528 & 121,082 & 97,446 & 10,209 \\ \hline
2) Normalization and Injectivity   & 156,146 & 109,961 & 46,185  & 9,203 \\ \hline
3) Disambiguation and Page Removal & 142,790 & 104,427 & 38,363  & 8,444 \\ \hline
4) Transitive Closure Removal      & 142,530 & 104,336 & 38,194  & 8,431 \\ \hline
5) Transitive Closure Addition     & 296,331 & 216,659 & 79,672 & 39,348 \\ \hline
6) Removal of exterior links     & 275,103 & 204,424 & 70,679 & 36,230 \\ \hline
\end{tabular}
\end{adjustbox}
\centering
\end{table}

In this section, the generated dataset is profiled and evaluated.
Table \ref{tab:eval} shows the corresponding number of extracted links for each step in the previous section. The reduction between the first and second steps is due to the normalization of directions and removal of links to ensure injectivity of the alignment. The fifth step does not only double the number of links but also increases the number of involved KG pairs by five, i.e., it adds a lot of links between graphs for which no links existed before.
Furthermore, we depict the number of trivial (i.e., sharing the same label) and non-trivial correspondences. In the final dataset, 25.7\% of all matches do not share the label and show that over a quarter of non-trivial cases (hard positives) are contained.

In the next step, the transitive closure is analyzed. 
Overall, 87,926 identity sets exist, consisting of $2.56 \pm 1.66$ entities on average. The largest identity set has 75 elements, referring to the entity \emph{Human}.

The final set of correspondences still contains links to web pages that are not part of Fandom and thus not available as a KG (step 6). Those links are kept until this point because they might also help to find correspondences under the transitive closure. After removing those links as well, we end up with 275,103 links to use. 
In the next step, the distribution of the confidence is analyzed. Out of the 132,961 links which are directly extracted (i.e., without exploiting transitivity), 125,407 have a confidence of 0.5 (only one direction) and 7,554 have links in both directions.
Overall 142,142 links could be generated by using the transitive property of the links. 
There is a huge amount of correspondences with a confidence of 0.4. This usually represents the situation where the correspondences A-B and B-C are extracted with a confidence of 0.5 each, and the link between A and C is created (no further path between A and C exists). Thus the maximum flow is 0.5 and the shortest path length is 2.0 (because an edge weight of 0.5 corresponds to a length of 1.0).

Evaluating the gold standard was done by a survey with three participants.
They were asked to judge if the sampled correspondences are correct.
Each group of correspondences is evaluated on its own: (1) extracted links with the confidence of 1.0, (2) extracted links with the confidence of 0.5, (3) transitive links with a confidence greater than 0.4, (4) transitive links with confidence equal to 0.4, and (5) transitive links with a confidence lower than 0.4. The last three groups are chosen this way because there are so many correspondences with a confidence of 0.4. The sample size is determined by a sample size calculator\footnote{\url{https://www.calculator.net/sample-size-calculator.html}}.
The confidence level is set to 90\% and a 10 \% margin of error with a population proportion of 50\% (which is very pessimistic because more correct correspondences are expected).
Table \ref{tab:survey} shows the sample sizes and the precision of the subset. The majority vote of the participants' answers is used to compute the final decision. The Fleiss kappa (inter-rater agreement) is 0.707 which is according to \cite{landis1977measurement} a substantial agreement (even though \cite{landis1977measurement} only provides this interpretation for two annotators, the value should at least be the same or better for three annotators).
The precision is always above 0.927 which shows the quality of the dataset. Depending on the application, it is further possible to increase it by using specific subsets or cutting the alignment at a specific confidence. The schema matches are extracted where $Sim_{Min}$ is greater than 0.2. The class matches are evaluated once with the class equivalence relation in mind, and once where the class can also be a subclass of the other (the value in parentheses). It shows that it is not easy to differentiate between \texttt{owl:\-equivalent\-Class} and \texttt{rdfs:\-subClassOf} during the automatic extraction.
For properties, this works much better, because a property match requires having matched subject \textit{and} object, whereas for classes, only sets of entities can be compared.

The limitation of this dataset is the class matches which should further be improved to distinguish between equivalence and subclass relations.

\begin{table}[t]
\centering
\caption{Evaluation of five subsets of the gold standard and the schema.}
\label{tab:survey}
\begin{tabular}{|l|r|r|r|}
\hline
                                     & \#All Links & \#Samples & Precision \\ \hline
(1) conf=1.0                         & 7,554       & 68        & 0.985          \\ \hline
(2) conf=0.5                         & 125,407     & 69        & 0.927          \\ \hline
(3) transitive conf \textgreater 0.4 & 8,098       & 68        & 0.985          \\ \hline
(4) transitive conf=0.4              & 109,064     & 69        & 0.941          \\ \hline
(5) transitive conf \textless 0.4    & 24,980      & 68        & 0.985          \\ \hline
classes ($Sim_{Min}>0.2$)    & 238      & 54        &  0.431 (0.922) \\ \hline
properties ($Sim_{Min}>0.2$)    & 786      & 63        &  0.921         \\ \hline
\end{tabular}
\end{table}

\section{Application of the Gold Standard}
\label{sec:application}
As discussed above, there exist many KG and ontology matching systems -- especially those which participate in the OAEI~\cite{oaei2021}. The main drawback is that all systems focus on a one-to-one alignment and not on the multi source case.
But it is possible to reuse these systems also for a multi source scenario.
The idea is to invoke the matchers multiple times to well-defined one-to-one matching tasks while avoiding complete quadratic comparisons.
Based on~\cite{ordermatters}, we implemented the incremental merge approach where two KGs are matched and merged to create a union out of them. Afterward, those are used in further match and merge steps. The execution order is determined by the output of a hierarchical agglomerative clustering (HAC), trying to match more closely related KGs first. The features for this clustering are tf-idf vectors generated from each KG. For this, all literals in the KG which contain text and all URI fragments are used and the concatenated text is transformed to tf-idf vectors. The feature space is restricted to only those words which appear at least in 0.1\% and at maximum in 80\% of the wikis.

The described multi source matching strategy is applied to the 40K dataset by using two different $1:1$ matchers. The first is a string-based based matcher, the second is Alod2Vec~\cite{alod2vec}, one of the top-performing systems from OAEI 2021~\cite{oaei2021} for the KG track. For both matchers, a postprocessing step is added to ensure a $1:1$ alignment which is required by the multi source matching strategy in each execution step. For this, a naive descending approach~\cite{mappingextraction} is used which sorts the alignment by confidence and extracts the correspondences as long as no duplicate source or target appears.

\begin{table}[t]
\centering
\caption{Evaluation of the multi source matching strategy applied to the 40K dataset reusing $1:1$ matchers (leftmost column). For each category, precision (P), recall (R), and f-measure (F) is reported.}
\label{tab:application}
\begin{adjustbox}{max width=\textwidth}
\begin{tabular}{|l|crr|crr|crr|crr|}
\hline
 &
  \multicolumn{3}{c|}{Instance} &
  \multicolumn{3}{c|}{Class} &
  \multicolumn{3}{c|}{Property} &
  \multicolumn{3}{c|}{Overall} \\ \hline
Matcher &
  \multicolumn{1}{c|}{P} &
  \multicolumn{1}{c|}{R} &
  \multicolumn{1}{c|}{F1} &
  \multicolumn{1}{c|}{P} &
  \multicolumn{1}{c|}{R} &
  \multicolumn{1}{c|}{F1} &
  \multicolumn{1}{c|}{P} &
  \multicolumn{1}{c|}{R} &
  \multicolumn{1}{c|}{F1} &
  \multicolumn{1}{c|}{P} &
  \multicolumn{1}{c|}{R} &
  \multicolumn{1}{c|}{F1} \\ \hline
Alod2vec &
  \multicolumn{1}{r|}{.937} &
  \multicolumn{1}{r|}{.390} &
  .551 &
  \multicolumn{1}{r|}{.842} &
  \multicolumn{1}{r|}{.432} &
  .571 &
  \multicolumn{1}{r|}{.000} &
  \multicolumn{1}{r|}{.000} &
  .000 &
  \multicolumn{1}{r|}{.936} &
  \multicolumn{1}{r|}{.390} &
  .550 \\ \hline
String-based &
  \multicolumn{1}{r|}{.959} &
  \multicolumn{1}{r|}{.679} &
  .795 &
  \multicolumn{1}{r|}{.829} &
  \multicolumn{1}{r|}{.392} &
  .532 &
  \multicolumn{1}{r|}{.702} &
  \multicolumn{1}{r|}{.549} &
  .617 &
  \multicolumn{1}{r|}{.959} &
  \multicolumn{1}{r|}{.679} &
  .795 \\ \hline
\end{tabular}
\end{adjustbox}
\end{table}

Table \ref{tab:application} shows the results for both matchers used in the multi source matching strategy. Similar to the results of OAEI, Alod2Vec achieves higher scores for class matches than the string-based matcher. For instance matches the string-based matcher is better. One reason might be that the Alod2vec matcher does not work well with the multi source strategy (which further requires the transitive closure during evaluation). The recall values show that there is still room for improvement and that the gold standard contains enough correspondences which are not easy to find.
The runtime of the two matchers are quite different. The string-based matcher needs roughly one day for merging and one day for matching. Alod2Vec needs overall six days most probably due to costly matching approaches.

\section{Conclusion and Outlook}
\label{sec:conclusion}

In this paper, we presented \emph{Gollum}, a large scale gold standard for multi source knowledge graph matching. The extraction process from more than 300,000 KGs is described and the resulting dataset is analyzed. It showed a high quality which can be further adjusted by the confidence depending on the application. In addition, all KGs and reasonable subsets of them are published as well as training and testing sets to account for systems adapting to the matching task.
To this end, we also created two different splits: one which contains links of the same wiki pair in train and test and one where each KG is either in the train or test split. It is further optimized that the test set cannot easily be computed by the transitive closure.
In future work, it is planned to improve the multi source matching strategy by using the training set to optimize the resulting alignment which is generated in each matching step. Thus one to one matchers would also benefit from the supervision even if the system is not aware of it.

\bibliography{Gollum}
\bibliographystyle{plainnat}

\end{document}